***Is Einstein more agreeable and less neurotic than Hitler? A computational exploration of the emotional and personality profiles of historical persons***

Arthur M. Jacobs[1,2] and Annette Kinder[3]

Author Note

1) Department of Experimental and Neurocognitive Psychology, Freie Universität Berlin, Germany

2) Center for Cognitive Neuroscience Berlin (CCNB), Berlin, Germany

3) Department of Education and Psychology, FUB, Germany

Correspondence: Arthur M. Jacobs

Department of Experimental and Neurocognitive Psychology, Freie Universität Berlin, Habelschwerdter Allee 45 , D-14195 Berlin, Germany.

Email: ajacobs@zedat.fu-berlin.de

## Abstract

Recent progress in distributed semantic models (DSM) offers new ways to estimate personality traits of both fictive and real people. In this exploratory study we applied an extended version of the algorithm developed in Jacobs (2019) to compute the likeability scores, emotional figure profiles and BIG5 personality traits for 100 historical persons from the arts, politics or science domains whose names are rather unique (e.g., Einstein, Kahlo, Picasso). We compared the results produced by static (word2vec) and dynamic (BERT) language model representations in four studies. The results show both the potential and limitations of such DSM-based computations of personality profiles and point ways to further develop this approach to become a useful tool in data science, psychology or computational and neurocognitive poetics (Jacobs, 2015).

*Sooner or later every psychologist working in the field of personality collides with the problem of trait-names.*
*Whatever method he employs,—rating scales, tests, factor-analysis, clinical interviews or any other technique,— he is*
*forced to ask himself whether the terms he is using in describing qualities and attributes of personality do actually*
*denote psychic dispositions or traits, or whether these terms are mischievous verbal snares tempting him into the*
*pitfalls of hypostaziation and other perils of 'verbal magic'*

— Gordon W. Allport and Henry S. Odbert, 'Trait-Names: A psycho-lexical study', 1936

## 1 Introduction

Carving the 'personality' of human beings into words is as old as human language itself and the
fields of personality, social and clinical psychology have been dominated by so-called lexical
approaches to personality assessment since the appearance of Galton's (1884) famous
'Measurement of Character' (cf. John et al., 1988). Such approaches are based on common
language descriptors and thus on the association between (mainly self-reported) words rather
than on observations. The language descriptors or 'trait-names', as Allport and Odbert (1936)
coined them, of the myriad of personality assessment tools to be found in psychology vary in
quantity and breadth (e.g., Hampson et al., 1987), but at the broadest level of abstraction this
domain still is most often represented by five dimensions usually called the 'BIG5' (Norman,
1963). Another term for this presumably most popular 'psycho-lexical' framework is the
*OCEAN* model, based on the initial letters of its five global factors (Openness to
experience/Intellect, Conscientiousness, Extraversion, Agreeableness, and
Neuroticism/Emotional Instability) assumed to capture personality characteristics that are most
important in people's lives. These are hypothesized to eventually become part of their language
and are more likely to be encoded into language as a single word than others. Given that the
number and interpretation of the terms used by the BIG5 model or its many alternatives are still
debated (e.g., Thompson, 2008), these five dimensions and their underlying 'lexical ground' (i.e.,
the word lists corresponding to each dimension) should not be considered final but as a
heuristically useful framework.

## 2 Related Work

Recent static and dynamic distributed semantic models (DSM) offer new ways to test at least the
face validity of psychological personality theories like BIG5 by estimating the 'digital personality
traits' of both fictive and real people. Thus, Egloff et al. (2016) applied the BIG5 model to
compute Hamlet's or Othello's personality profiles. In a more recent and extensive study, Jacobs
(2019) computed the emotional and BIG5 personality profiles of 100 characters from the popular
'Harry Potter' book series (e.g., Rowling, 1997). Using the computational "goodness/badness" of character for the 100 figures as a superordinate class label, Jacobs cross-validated his results on the basis of Harry Potter homepages that categorized them—as clearly as possible—as either "good" (e.g., "friend of Harry," "the Weasleys") or "bad" (e.g., "enemy of Harry," "death eaters"). The data showed a classification accuracy of just over 80% which was considered encouraging for pursuing the approach.

## 3 Models and Methods

Here we applied an extended version of the algorithm developed in Jacobs (2019) to compute the likeability, emotion profiles and BIG5 personality traits for 100 historical persons from different domains (art, politics and science). The list given in Appendix A is idiosyncratic, i.e. a subjective selection of the authors with a focus on famous historical persons from three domains whose names are rather unique and thus less likely to confuse with other persons or terms. Our aim was to examine to what extent and under which conditions Jacobs' method could also be applied to real persons referred to in 'wikipedia' (i.e., being part of the vocabulary of DSMs trained on wikipedia pages). In particular, we tested whether static and dynamic language model representations produced congruent results. In short, static models like word2vec (Mikolov et al., 2013) produce a single vector per word and thus a practically context independent representation. In contrast, dynamic models like BERT (Devlin et al., 2019) can have several context dependent representations for a given word. It was interesting to see which of these models produced results with greater face validity. The two DSMs were downloaded from: static ('wiki.en.vec'; https://fasttext.cc/docs/en/pretrained-vectors.html) and dynamic (https://huggingface.co/sentence-transformers/bert-base-nli-mean-tokens).

## 4 Experiments

We tested the face and empirical validity of our models in four experiments. In study 1 we compared the likeability scores computed for the 100 historical persons by our two models using an empirically well validated word list (Anderson, 1968). In studies 2 and 3 we compute the emotional figure profiles and test the empirical validity of the likeability and emotional figure profile scores in predicting human liking ratings using an extended database of N>13k as ground truth (Warriner et al., 2013). In the final study 4 we compare both models on their BIG5 computations for the 100 persons.

### Study 1. Likeability

A widely used tool for assessing the likeability of persons is Anderson's (1968) list of 550 words assumed to indicate to which extent a person is more likeable than another. Here we used the top 100 words from that list, i.e. those with the highest likeability ranks, as well as the 'flop 100' (i.e. those with the highest dislikeability ranks; see Appendix B) to compute a likeability score (i.e., the difference between the means for the likeability and dislikeability values) for each of the 100 persons. Figure 1 summarizes the results showing the likeability score (average likeability value – average dislikeability value) for the 100 persons. The results appear to bear some face validity as far as we can tell, although there are clear discrepancies between the values based on the static model (w2v) and those for the dynamic model (BERT). Although the overall correlation between the two measures is weak (r = .11), it is encouraging to see that they both see the nobelist Lise Meitner and the singer Caruso as extremely likeable persons and Hitler as the most dislikeable. Notable disagreement between the models appears for the German poet Goethe, who is likeable only according to w2v, or the actor Marlon Brando who is likeable only according to BERT, and curiously aside of the 'nazi' Eichmann.

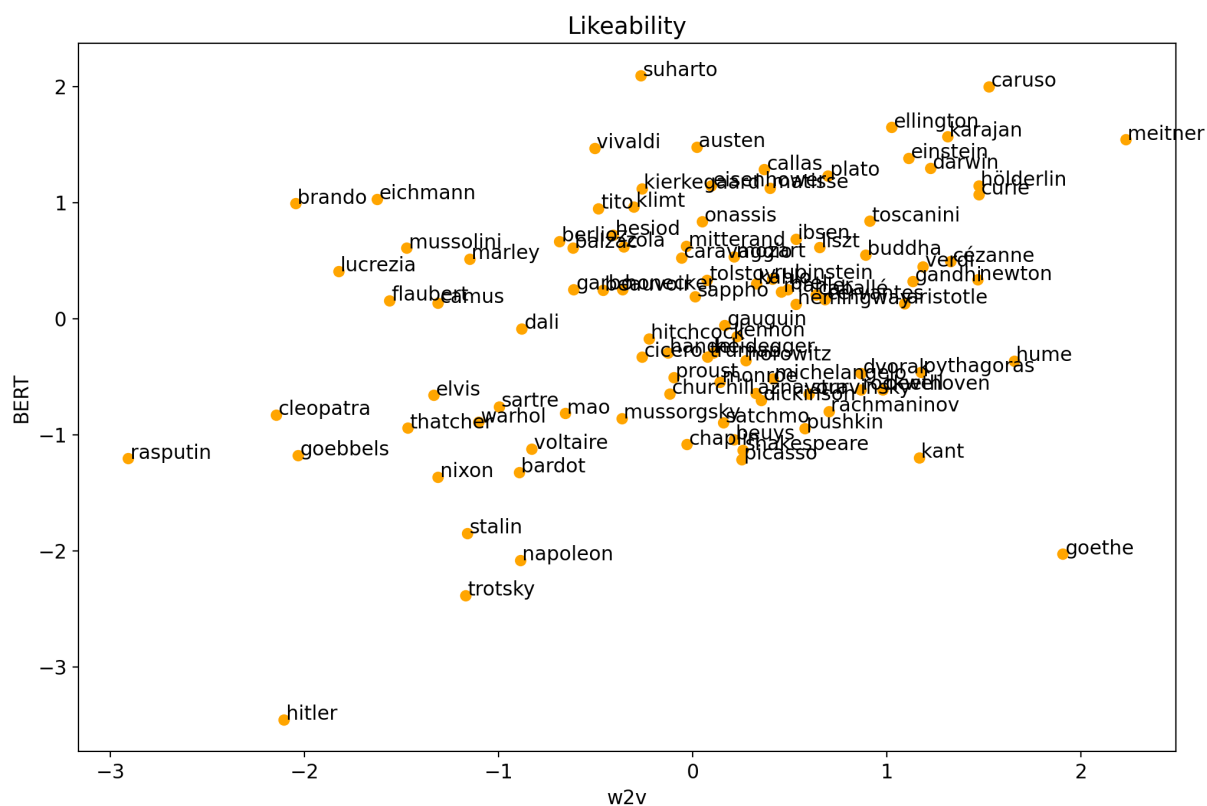

**Figure 1. Likeability Scores (z-values) for 100 persons based on the w2v and BERT models**

For lack of any 'gold standard' or ground truth and in order not to risk hurting any reader's feelings we will refrain from interpretations of our own regarding the subjective or general plausibility of these data and leave that task to readers. It should be kept in mind that these data do not reflect 'true' personality traits or feelings about these persons, but only computational estimates of associations between names and pieces of information sampled on a particular day from wikipedia, as learned by two types of neural net models. Just as real 'psycho-lexical' personality assessments of either state or trait features, they can be more or less close to reality. To establish the empirical validity of such data would require extensive and costly studies using living persons and trained personality psychologists. Although this is possible in principle, it was not the goal of this paper. Explaining the discrepancies between the two models remains a challenge for future studies closely looking at the underlying training corpora or their hyper-parameters.

In sum, the results of study 1 suggest that when applied to the 200 extreme words of Anderson's (1968) likeability list both static and dynamic DSMs produce partly comprehensible results (e.g., Meitner or Hitler), as well as surprises (e.g., Goethe, Brando or Eichmann). Since we did no hyperparameter or fine tuning for either model, we will not speculate to what extent the present results are generalisable, though. Using other w2v or BERT models with different training corpora or hyperparameter sets may produce quite different results. We leave this task to future papers, the aim of the present one being to check the general feasibility of our computational approach to digital personality assessment. To complement and cross-validate the likeability list data, in the next study we computed the emotional figure profile (EFP) for our 100 person set.

## Study 2. Emotional Figure Profile (EFP)

The computation of the EFP of a fictive or real character is based on the most widely applied emotion theory of psychology (Wundt, 1874) which locates all emotions in a 2D space spanned by the dimensions valence (negative to positive) and arousal (calming to arousing). Thus, the emotion potential of a single word is computed as: $EP_w = |valence_w| * arousal_w$, and estimates the bivariate potential with which a word or larger text unit (i.e., average across all words in that unit) can elicit emotional responses in readers, independently of the valence sign. Since even after more than 150 of scientific psychology there still is no consensus which of the two major emotion theories is correct (cf. Schrott and Jacobs, 2011): dimensional theories of emotion (e.g., Wundt's valence and arousal) or discrete ones (e.g., Ekman, 1999). The DSM-based variant of the emotion potential proposed by Jacobs (2019) takes both theoretical approaches into account: its computation is based on discrete labels (e.g., *joy*, *fear* etc.; see Appendix 2 in Jacobs, 2018), but its output is a continuous value on the bipolar "negativity-positivity" and "calming/arousing" dimensions, respectively.

The EFP thus comprises the valence, arousal and emotion potential values of a character, i.e. the extent to which this character can arouse emotions in readers. Jacobs (2019) computed the EFPs for 100 characters from the Harry Potter series and showed that among seven major figures *Voldemort* had the highest emotion potential due to his extreme arousal value (percentile 99), while *Hermione* also had quite a high emotion potential (percentile 82) but due to her very high valence value (percentile 88).

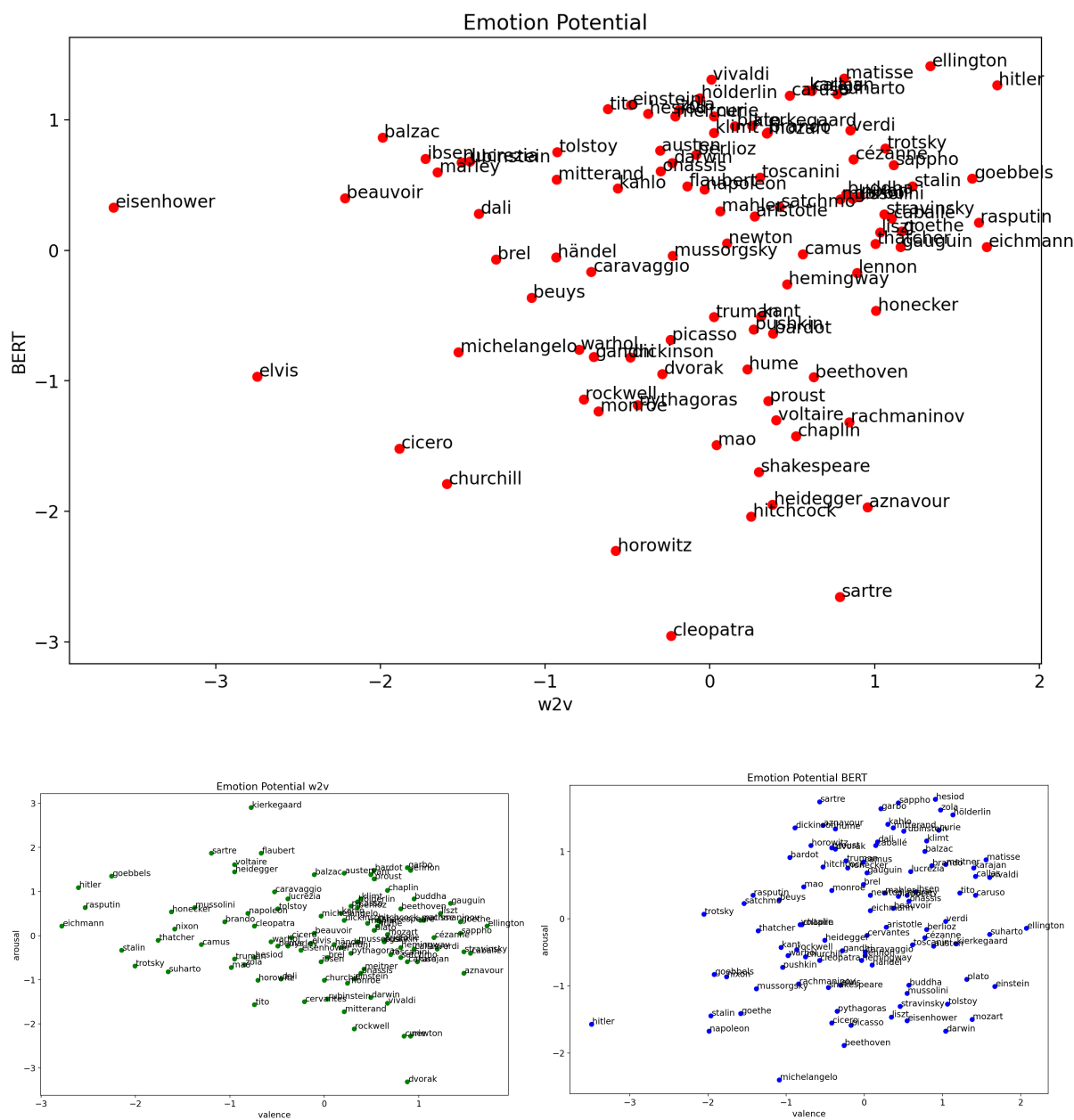

Figure 2. Emotion Potential (a) and Valence and Arousal Scores (b+c) for 100 persons based on the w2v and BERT models.

The data in Figure 2[1] again show both similarities and differences between the two DSMs. Thus, both models agree in Figure 2a that Hitler and Duke Ellington have the highest emotion potential, albeit for different reasons. This can be seen in Figures 2b and c which break down the overall scores of Figure 2a into their valence and arousal components showing Hitler's high w2v arousal score (Figure 2b) and his (negative) w2v and BERT valence scores (Figure 2c), as well as Duke Elligton's top (positive) valence scores in both models. On the other hand, there also is considerable disagreement between the models: for example, the w2v model estimates that Sartre has a very high emotion potential, while BERT assigns a very high score to Eisenhower. Overall the correlation between the emotion potential scores of the two models is very low (r = .1).

Regarding correlations with the likeability scores of study 1, they were r = .74 for w2v and r = .94 for BERT valence scores. This lends support to the idea that all three methods can in principle be used to computationally operationalize the psychological construct of 'liking' which has been shown to be a very good predictor of aesthetic feelings for both pictorial and verbal materials (for review, see Jacobs et al., 2016).

To sum up this section, just like likeability scores the results of EFP computations differ considerably between the two types of DSMs. While some results could be said to have face validity in both models (e.g., Hitler's very negative vs. Ellington's very positive valence scores) there are also many puzzling results (e.g., Kierkegaard's and Sartre's extreme arousal values in w2v vs. BERT, respectively) which require further examination in future studies using different training corpora or model hyperparameters. Finally, study 1's likeability scores correlated very highly with the present valence values for both models raising the question which method has higher empirical validity, i.e. yields better fits with human valence ratings. Study 3 examines this issue.

## Study 3. Empirical Validation of Likeability and Valence Score Methods

For the computational data of the first two studies there is no such thing as a ground truth leaving plausibility considerations untested against reality. In study 3 we therefore used the most extensive English database on word valence (N > 13k words; Warriner et al., 2013) as ground truth for examining the empirical validity of our DSMs. The results of study 3 shown in Figure 3 show very good fits for both models when valence/AAPz was the predictor (with a clear advantage for the BERT models). This establishes the present w2v and –for the first time also

---

[1] The data in Figure 2 are all standardized (z-values) and those in Figure 2a give the log values to compensate for their non-normal distribution.

BERT– as valid models of human valence ratings achieving similarly high correlations as previous studies using different DSMs (e.g., Hollis et al., 2016; Jacobs & Kinder, 2019).

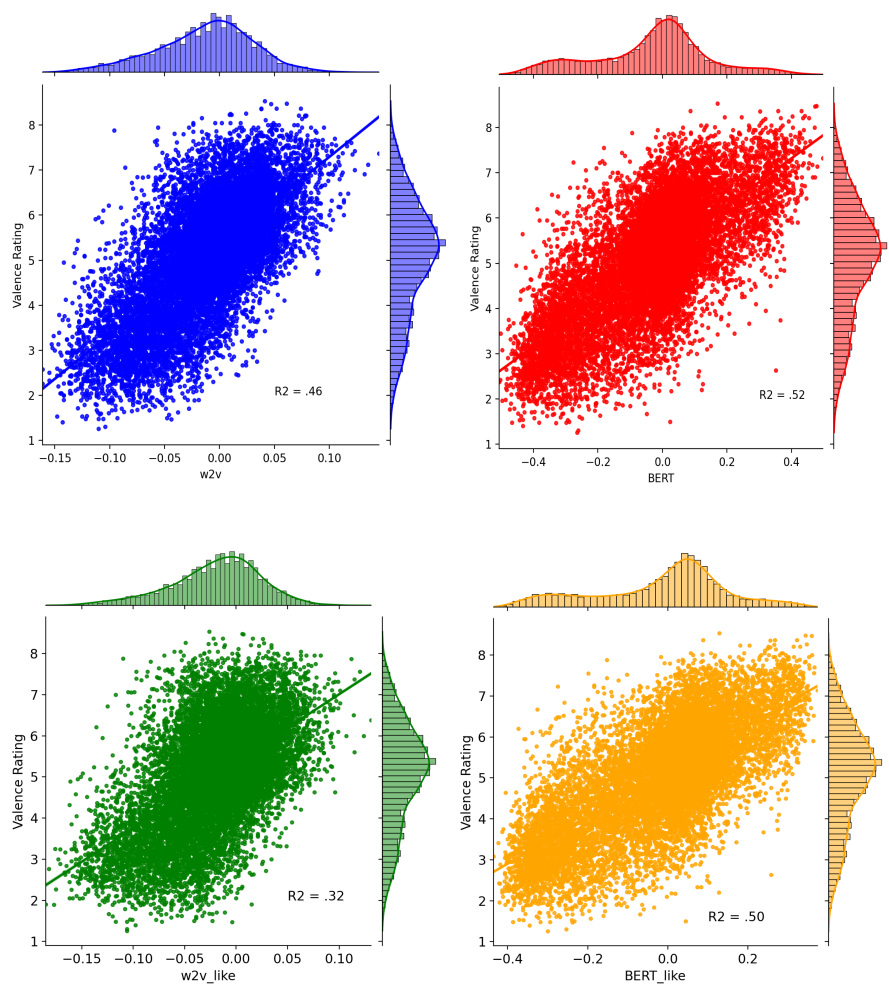

**Figure 3. Correlations between human valence ratings from Warriner et al. (2013) and w2v vs BERT valence (upper panel) and likeability (lower panel) scores**

## Study 4. BIG5 Mini-markers

The results of Jacobs (2019) regarding the BIG5 personality profile (based on a variant of the OCEAN model) of seven major figures from Harry Potter bore sufficient face validity to further pursue this approach. Thus, quite plausibly while Voldemort featured the highest score on the N dimension (Neuroticism, Emotional Instability), Harry scored highest on C, O and A (Conscientiousness, Openness/Intellect and Agreeableness). Extending our earlier work, here we chose to use a well established tool for measuring the BIG5, Thompson's (2008) 'International

English Big-Five Mini-Markers'. With 40 items[2] (~8 per dimension), this tool represents a good compromise between tools like the BFI10 which uses only 10 items (2 per dimension) to assess a 'personality' (Rammstedt & John, 2007) and extensive but hard to handle tools like Norman's (1968) list of 2800 words.

For each of the 100 persons[3] we computed the association with the terms corresponding to the positive and negative poles of each of the five dimensions (e.g., the cosine similarity between the vector for 'Einstein' and the vectors for 'uncreative' and 'unimaginative' representing the negative pole of the Openness dimension). The average values across all positive and negative terms were then subtracted from each other to obtain the final scores (see Appendix D for an example code). We also averaged the values for both models in an attempt to obtain more objective and potentially valid results.

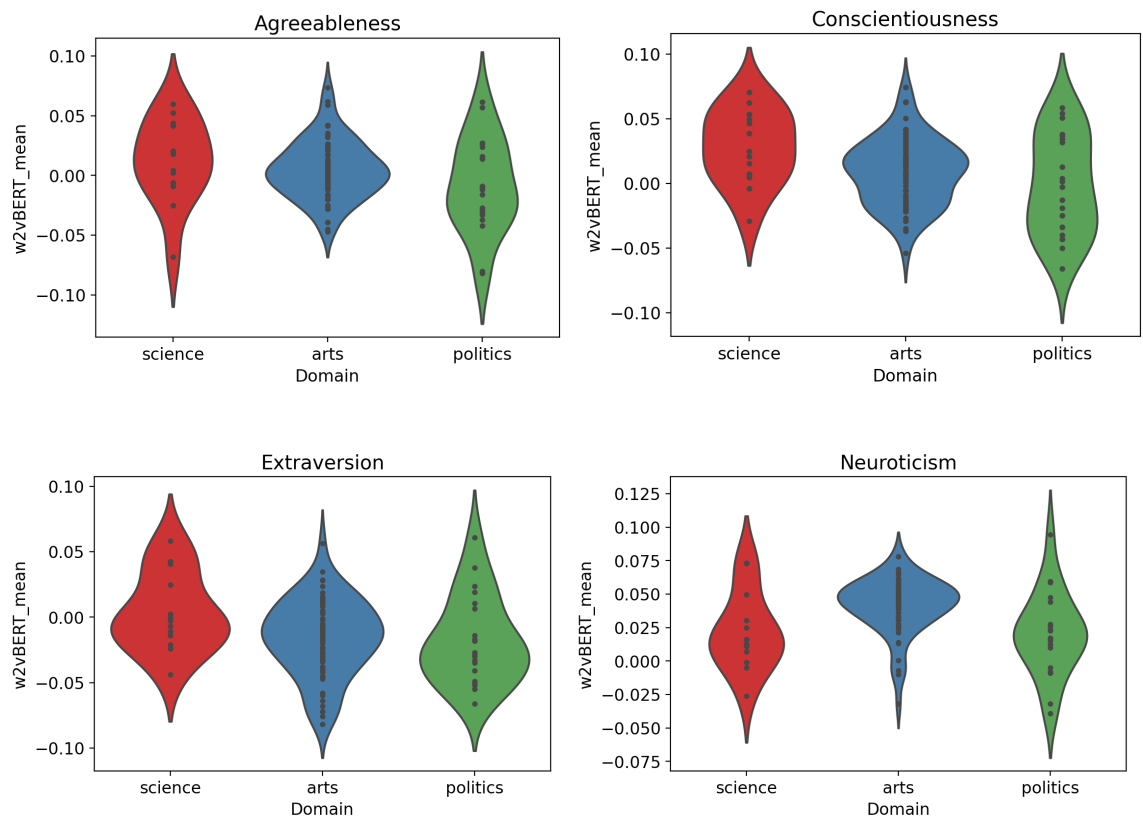

---

Running head: Computing the emotional and personality profiles of historical persons

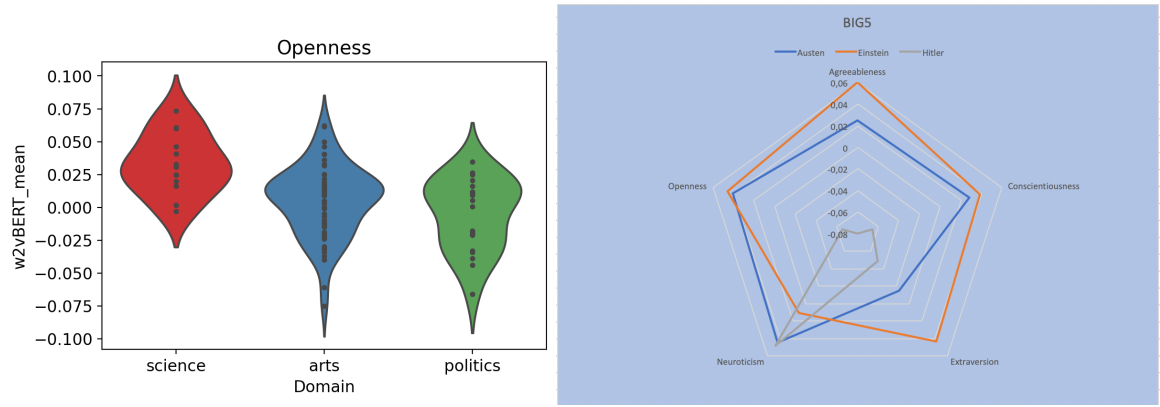

**Figure 4. a-e (violin plots) BIG5 scores for three domains based on average of w2v and BERT models, f (radar plot) BIG5 scores for three persons representative of the three domains**

The data in Figure 4a-e (violin plots) suggest that at least for the present sample of 13 scientists these persons score highest in all but one BIG5 dimension: in order of effect size, they clearly show the highest average Openness/Intellect (0.03 > .005 (arts) > -.005 (politics); p <.0003, $R^2$ = .15), Conscientiousness (0.03 > .001 > -.002; p <.0003, $R^2$ = .08), Extraversion (0.003 > -.016 > -.018; p <.088, $R^2$ = .05) and Agreeableness scores (0.012 > .005 > -.001; p <.094, $R^2$ = .048), while our sample of 64 artists prevails on the Neuroticism dimension, the 21 politicians appearing to be the most emotionally stable: .04 > .022 (science) > .02 (politics), p <.0005, $R^2$ = .146. The exemplary data in Figure 4f (radar plot) give a response to the question asked in this paper's title and display Einstein as the top scorer on 4 of the BIG5 dimensions followed by Jane Austen. Hitler only leads on the Neuroticism dimension, closely followed by Austen. Thus, if one believed in the validity of the present models, Einstein indeed was clearly more agreeable and less neurotic than Hitler.

## 5 Discussion, Limitations and Outlook

The present exploratory studies show the potential and limitations of a DSM-based computational approach to personality assessment. While a previous study on fictive persons from the Harry Potter book series (e.g., Rowling, 1997) provided encouragingly plausible results (Jacobs, 2019), the data of these four studies suggest that there is still a lot of work to do in this novel field of research. Whether the likeability, emotion potential or BIG5 scores are considered, our two DSMs conflict more than they agree in their assessment. Where they agree, the results seem not implausible, but there is too much divergence to allow any conclusions at this stage. Still we see enough potential in this approach to encourage future work that focuses on hybrid models combining static and dynamic DSMs (e.g., by concatening vectors from w2v and BERT models) using the same training corpus and extensive hyperparameter tuning studies, and, ultimately, a ground truth provided by human ratings of the personality traits of the analyzed fictive or real persons.

In any case our approach can be used to examine the validity of pretrained DSMs that are publically available on the internet (e.g., fasttext or hugging face) for assessing the 'digital personality' of fictive, historical or living persons. Thus, standard quality tests for DSMs such as solving analogies like 'King is to man as X is to woman' or the 'which one doesn't fit' problem (France, Germany, Italy, Africa) could be augmented by testing to what extent different pretrained models produce plausible associations between items used in NER lists (e.g., names of persons, institutions etc.) and standard terms used by psychologists to assess emotional and personality profiles.

One key task for such future work in data science and NLP is the development of appropriate methods for disambiguation, the uniqueness vs. ambiguity ('confusability') of e.g. person names being a non trivial issue (e.g., the name 'Onassis' in our list may refer to either the Greek millionaire or his wife). Another issue concerns the appropriateness and validity of word lists provided by psychological personality theories such as those used here. About 150 years after Galton's (1884) 'Measurement of character' personality psychologists are still far from a general consensus on exactly how many dimensions are necessary or sufficient to scientifically describe a personality and how many items or words are necessary or sufficient to assess these dimensions. In short, not only the computational but also the psychological side of the present approach are work in progress and we believe that both sides can benefit from scientific cross-fertilization. To paraphrase the initial quote of this paper, despite the mixed results we obtained with our two models, together with their original authors we hope that the word lists explored here are more than '*mischievous verbal snares tempting us into the pitfalls of hypostaziation and other perils of verbal magic*'.

## Appendix

### A. 100 person names

pers =
['aristotle','austen','aznavour','balzac','bardot','beauvoir','beethoven','berlioz','beuys','brando','brel',
'buddha','caballé','callas','camus','caravaggio','caruso','cervantes','cézanne','chaplin','churchill',
'cicero','cleopatra','curie','dali','darwin','dickinson','dvorak','eichmann','einstein','eisenhower',
'ellington','elvis','garbo','flaubert','gandhi','gauguin','goebbels','goethe','händel','heidegger',
'hemingway','hesiod','hitchcock','hitler','hölderlin','honecker','horowitz','hume','ibsen','kahlo',
'karajan','kant','kierkegaard','klimt','lennon','liszt','lucrezia','mahler','mao','marley','matisse','meitner',
'michelangelo','mitterand','monroe','mozart','mussolini','mussorgsky','napoleon','newton','nixon',
'onassis','picasso','plato','proust','pushkin','pythagoras','rachmaninov','rasputin','rockwell',
'rubinstein','sappho','sartre','satchmo','shakespeare','stalin','stravinsky','suharto','thatcher','tito',
'tolstoy','toscanini','trotsky','truman','verdi','vivaldi','voltaire','warhol','zola']

### B. 200 likeability terms (cf. Anderson, 1968)

positive = [
 'sincere',
 'honest',
 'understanding',
 'loyal',
 'truthful',
 'trustworthy',
 'intelligent',
 'dependable',
 'wise',
 'considerate',
 'reliable',
 'mature',
 'warm',
 'earnest',
 'kind',
 'friendly',
 'happy',
 'clean',
 'interesting',
 'unselfish',
 'honorable',
 'humorous',
 'responsible',
 'cheerful',
 'trustful',
 'gentle',

'educated',
'reasonable',
'companionable',
'likable',
'trusting',
'clever',
'courteous',
'tactful',
'helpful',
'appreciative',
'imaginative',
'outstanding',
'brilliant',
'enthusiastic',
'polite',
'original',
'smart',
'forgiving',
'ambitious',
'bright',
'respectful',
'efficient',
'grateful',
'conscientious',
'resourceful',
'alert',
'good',
'witty',
'kindly',
'admirable',
'patient',
'talented',
'perceptive',
'spirited',
'sportsmanlike',
'cooperative',
'ethical',
'intellectual',
'versatile',
'capable',
'courageous',
'constructive',
'productive',
'progressive',
'individualistic',
'observant',
'ingenious',
'lively',
'neat',
'punctual',
'logical',
'prompt',

'accurate',
'sensible',
'creative',
'tolerant',
'amusing',
'generous',
'sympathetic',
'energetic',
'tender',
'active',
'independent',
'respectable',
'inventive',
'wholesome',
'congenial',
'cordial',
'experienced',
'attentive',
'cultured',
'frank',
'purposeful',
'decent' ]

negative = ['messy',
'misfit',
'uninteresting',
'scornful',
'antisocial',
'irritable',
'stingy',
'tactless',
'negligent',
'weak',
'profane',
'gloomy',
'helpless',
'disagreeable',
'touchy',
'irrational',
'tiresome',
'disobedient',
'complaining',
'lifeless',
'vain',
'lazy',
'unappreciative',
'maladjusted',
'aimless',
'boastful',
'dull',
'gossipy',
'unappealing',

'hypochondriac',
'irritating',
'petty',
'shallow',
'deceptive',
'grouchy',
'egotistical',
'meddlesome',
'uncivil',
'cold',
'unsportsmanlike',
'bossy',
'unpleasing',
'cowardly',
'discourteous',
'incompetent',
'childish',
'superficial',
'ungrateful',
'unfair',
'irresponsible',
'prejudiced',
'bragging',
'jealous',
'unpleasant',
'unreliable',
'impolite',
'crude',
'nosey',
'humorless',
'quarrelsome',
'abusive',
'distrustful',
'intolerant',
'unforgiving',
'boring',
'unethical',
'unreasonable',
'snobbish',
'unkindly',
'unfriendly',
'hostile',
'dislikable',
'offensive',
'belligerent',
'underhanded',
'annoying',
'disrespectful',
'selfish',
'vulgar',
'heartless',
'insolent',

Running head: Computing the emotional and personality profiles of historical persons

```
 'thoughtless',
 'rude',
 'conceited',
 'greedy',
 'spiteful',
 'insulting',
 'insincere',
 'unkind',
 'untrustworthy',
 'deceitful',
 'dishonorable',
 'malicious',
 'obnoxious',
 'untruthful',
 'dishonest',
 'cruel',
 'mean',
 'phony',
 'liar']
```

## C. 40 Mini Markers (cf. Thompson, 2008)

**Openness/Intellect**. negative = ['uncreative','unimaginative']; positive = ['creative','intellectual','intelligent' 'philosophical','deep','artistic']

**Conscientiousness**. negative = ['disorganized','untidy','inefficient','careless']; positive = ['organized','neat','systematic','efficient']

**Extraversion**. negative = ['untalkative','quiet','shy','reserved']; positive = ['talkative','outgoing','energetic','efficient']

**Agreeableness**. negative = ['unkind','rude','inconsiderate','harsh']; positive = ['kind','sympathetic','warm'] #'cooperative' ist besser in Np

**Neuroticism/Emotional Instability**. negative = ['anxious','jealous','moody','emotional','envious']; positive = ['unworried','cooperative','resilient','stable']

## D. Example .py code: Computing the Openness score for 'Einstein' (using gensim library, https://radimrehurek.com/gensim/)

```python
#get libraries and DSM
from gensim.models import KeyedVectors
import numpy as np
TC = 'wiki.en.vec' # https://fasttext.cc/docs/en/pretrained-vectors.html
```

```
w2v_model = KeyedVectors.load_word2vec_format(TC,unicode_errors='ignore')
vocab = list(w2v_model.index_to_key)
print(len(vocab)) # 2.519.370

#get w2v similarity between all negative and positive terms in the above Openness/Intellect lists,
compute difference between mean values
negative = ['uncreative','unimaginative']; positive = ['creative','intellectual','intelligent'
'philosophical','deep','artistic']
p = 'Einstein'
On = [w2v_model.similarity(p,w) for w in negative]
Op = [w2v_model.similarity(p,w) for w in positive]
Open = np.mean(Op) - np.mean(On) #
```